\documentclass[letterpaper, 10 pt, conference]{ieeeconf}
\IEEEoverridecommandlockouts
\usepackage{cite}
\usepackage{amsmath,amssymb,amsfonts}
\usepackage{algorithmic}
\usepackage{graphicx}
\usepackage{textcomp}
\usepackage{color, soul}
\usepackage{xcolor}
\def\BibTeX{{\rm B\kern-.05em{\sc i\kern-.025em b}\kern-.08em
    T\kern-.1667em\lower.7ex\hbox{E}\kern-.125emX}}

\usepackage{makecell}
\usepackage{graphicx}
\usepackage{textcomp}
\usepackage{array} 
\usepackage{longtable}
\usepackage{booktabs}
\usepackage{float}
\usepackage{dblfloatfix}

\title{\LARGE \bf
Color-Pair Guided Robust Zero-Shot 6D Pose Estimation and Tracking of Cluttered Objects on Edge Devices
}

\author{Xingjian Yang$^{1}$ and Ashis G. Banerjee$^{2}$%
\thanks{$^{1}$X. Yang is with the Department of Mechanical Engineering, University of Washington, Seattle, WA 98195, USA.
{\tt\small yxj1995@uw.edu}}%
\thanks{$^{2}$A. G. Banerjee is with the Department of Industrial \& Systems Engineering and the Department of Mechanical Engineering, University of Washington, Seattle, WA 98195, USA.
{\tt\small ashisb@uw.edu}}
}

\begin{document}

\maketitle
\thispagestyle{empty}
\pagestyle{empty}

\begin{abstract}
Robust 6D pose estimation of novel objects under challenging illumination remains a significant challenge, often requiring a trade-off between accurate initial pose estimation and efficient real-time tracking. We present a unified framework explicitly designed for efficient execution on edge devices, which synergizes a robust initial estimation module with a fast motion-based tracker. The key to our approach is a shared, lighting-invariant color-pair feature representation that forms a consistent foundation for both stages. For initial estimation, this feature facilitates robust registration between the live RGB-D view and the object's 3D mesh. For tracking, the same feature logic validates temporal correspondences, enabling a lightweight model to reliably regress the object's motion. Extensive experiments on benchmark datasets demonstrate that our integrated approach is both effective and robust, providing competitive pose estimation accuracy while maintaining high-fidelity tracking even through abrupt pose changes.
\end{abstract}

\section{Introduction}
Estimation of an object's six-degree-of-freedom (6D) pose, which involves determining its 3D rotation and 3D translation relative to a camera, is a fundamental task in computer vision and robotics \cite{hodan2024bop}. Accurate 6D pose information is crucial for a variety of applications, ranging from robotic manipulation and grasping in industrial and household environments to immersive experiences in augmented and mixed reality. The ability of an autonomous system to precisely locate and determine the orientation of objects is a key prerequisite for meaningful physical interaction. Furthermore, in dynamic scenarios, this capability must extend beyond single-frame estimation to continuous, real-time tracking, providing the temporal coherence necessary for tasks such as closed-loop robotic control.

Historically, pose estimation has focused on instance-level methods, which require costly, object-specific training and thus cannot generalize to new objects. While category-level approaches can handle unseen instances within a known class, they still fail to address entirely novel categories. This fundamental generalization gap has motivated the shift to zero-shot pose estimation, which aims to handle any unseen object given its 3D CAD model.

    \begin{figure}[thpb]
      \centering
      \includegraphics[scale=0.18]{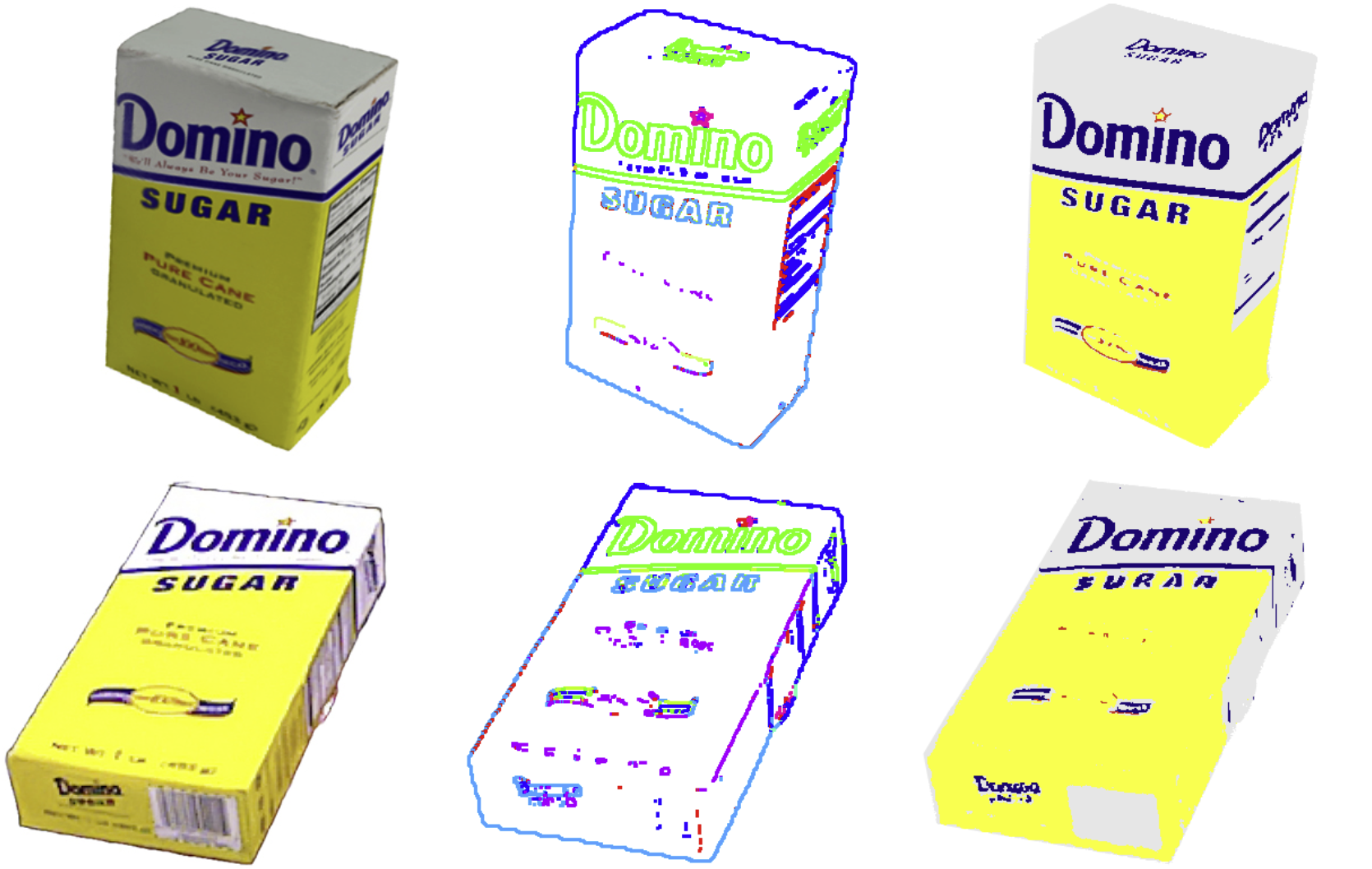}
      \caption{Visualization of color-pair based classification. Left to right: input image, color-pairs visualized by class, and classified surface patches based on matching. Rows compare results on synthetic (rendered) vs. real (captured) images under varying illumination to demonstrate robustness.}
      \label{main_concept}
   \end{figure}

    \begin{figure*}[!b]
      \centering
      \includegraphics[scale=0.2]{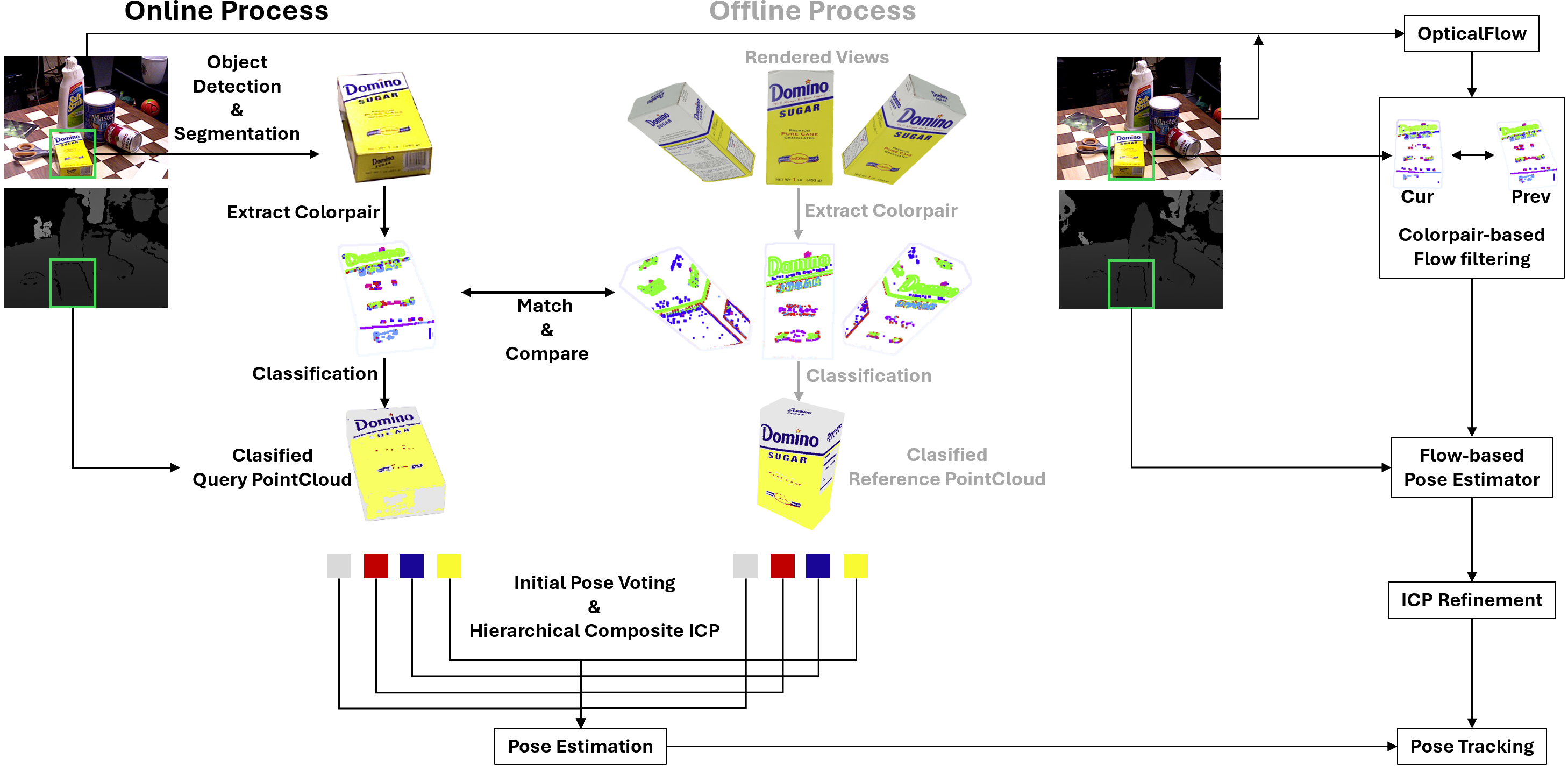}
      \caption{Overview of the pipeline. Pose estimation: After object detection and segmentation, color-pair features are extracted from the input RGB-D frame. These are matched against features from offline rendered views to generate a classified point cloud, which is then used in a pose voting and hierarchical ICP stage to determine the pose. Pose tracking: The pose is updated frame-to-frame by feeding color-pair filtered optical flow into a motion estimator that predicts the inter-frame transformation.}
      \label{pipeline}
   \end{figure*}
   
Recent advances in zero-shot pose estimation are largely driven by Vision Foundation Models (VFMs) such as DINOv2 \cite{oquab2023dinov2} and CLIP \cite{radford2021learning}. These models provide powerful descriptors that enable reliable correspondence matching between an input image and rendered templates, bridging the synthetic-to-real domain gap without task-specific fine-tuning. However, this feature matching paradigm often relies on exhaustive comparisons against a vast set of templates, resulting in significant latency. This computational cost is a critical bottleneck for real-time tracking, where systems must be resilient to drift and recover from occlusions without frequent, costly re-initialization. Further, establishing robust correspondences remains difficult for challenging objects, such as those that are texture-less or symmetric. Concurrently, emerging research explores generative models, such as Diffusion \cite{von2024diffusion}, as well as vision-language models (VLMs) \cite{corsetti2024open}, to address these limitations.

In this paper, we present a unified framework for 6D object pose estimation and tracking, specifically designed for textured objects under challenging real-world lighting conditions. Our approach integrates a novel, lighting-invariant feature representation for robust initial localization with a fast, motion-based model for real-time tracking in subsequent frames. Our main contributions are as follows:
\begin{itemize}
  \item A novel, lighting-invariant colorpair feature descriptor that unifies the framework by enabling both robust initial pose estimation and reliable correspondence filtering for tracking (see Fig.~\ref{main_concept})
  \item A robust and efficient initial pose estimation algorithm that leverages the spatial distribution of local texture features to register classified edge point clouds from a segmented RGB-D image against a 3D model
    \item An efficient 6D pose tracking module that leverages optical flow and depth cues with a viewpoint-invariant rotation estimator to achieve strong robustness against abrupt pose changes
\end{itemize}

\section{Related Works}
Contemporary zero-shot pose estimation approaches typically leverage 3D models and are distinguished by their methodology: some train on large, diverse datasets to learn generalizable features; others build on pre-trained foundation models for training-free inference; and still others explore alternative problem formulations.

\textbf{Training-based generalization for unseen objects.} A key line of work trains on large, diverse (often synthetic) datasets to generalize to novel objects. MegaPose \cite{labbe2022megapose} employs an effective render-and-compare paradigm that iteratively refines a coarse pose by comparing the observed image with renderings of the CAD model. FoundationPose \cite{wen2024foundationpose} unifies pose estimation and tracking within one framework trained on large-scale synthetic data. GenFlow \cite{moon2024genflow} learns to predict optical flow between rendered and observed images and refines pose iteratively under 3D shape guidance. Other approaches include GigaPose \cite{nguyen2024gigapose}, a retrieval-style orientation pipeline based on fusing local similarities.

\textbf{Training-free estimation with foundation models.} Another direction avoids task-specific training by leveraging frozen 2D/3D foundation models to establish reliable correspondences. RGB-D methods like FreeZeV2 \cite{caraffa2025accurate} combine vision-foundation patch features with geometric priors to construct 3D descriptors for 3D–3D registration. ZeroPose \cite{chen2024zeropose} adopts a related correspondence-driven strategy, while SAM-6D \cite{lin2024sam} uses SAM for object masks before matching the masked cloud to rendered views. In the RGB-only setting, FoundPose \cite{ornek2024foundpose} shows that intermediate ViT patch descriptors are sufficient for matching real images to pre-rendered CAD templates, enabling PnP/RANSAC pipelines without task-specific training. ZS6D \cite{ausserlechner2024zs6d} follows this correspondence or template-matching approach with frozen ViT features. Diffusion-backbone features have also been explored as competitive alternatives \cite{von2024diffusion}.

\textbf{Alternative representations and model-free formulations.} Beyond correspondence and render-and-compare pipelines, several works explore alternative representations or problem formulations. NOPE \cite{nguyen2024nope} and Zero123-6D \cite{di2024zero123} use image-to-3D view synthesis to generate pseudo-templates from a few reference images, reducing reliance on explicit CAD models. Any6D \cite{lee2025any6d} first reconstructs a 3D shape, then jointly refines its scale and pose. Other representations include graph-based methods like 3DPoseLite \cite{dani20213dposelite} and 3D Gaussian Splatting in GS2Pose \cite{mei2024gs2pose} for differentiable refinement. Finally, some works reframe the problem, using VLMs to locate objects by text \cite{pulli2024words} or formulating refinement as an action-decision process \cite{busam2020like}.

\textbf{6D Object Tracking.} 
While the above methods focus on single-frame estimation, a parallel line of work addresses 6D object tracking by modeling inter-frame motion and consistency for more efficient and stable pose updates. A classic paradigm is probabilistic filtering, with methods such as PoseRBPF \cite{deng2021poserbpf} using a particle filter to maintain a posterior over object orientations. Others, like ROFT \cite{piga2021roft}, employ Kalman filters to fuse cues such as optical flow for tracking fast-moving objects. More recently, learning-based approaches have become dominant, training networks to directly regress the relative transformation between frames (e.g., SE(3)-TrackNet \cite{wen2020se}) or to track a sparse set of keypoints. Methods like 6-PACK \cite{wang20206} learn category-specific keypoints, while model-free approaches such as BundleTrack \cite{wen2021bundletrack} track general features and use pose graph optimization for long-term consistency.

\section{Methods}

Given an RGB-D image of a textured object, our goal is to determine its full 6D pose and subsequently track its motion in a video sequence. Effectively tackling this problem requires a synergistic approach that leverages illumination-invariant texture features for robust initial localization and efficient motion cues for real-time tracking. Our unified framework, depicted in Fig. \ref{pipeline}, achieves this by integrating a novel pose estimation pipeline with a lightweight tracking module.

    \begin{figure}[b]
      \centering
      \includegraphics[scale=0.24]{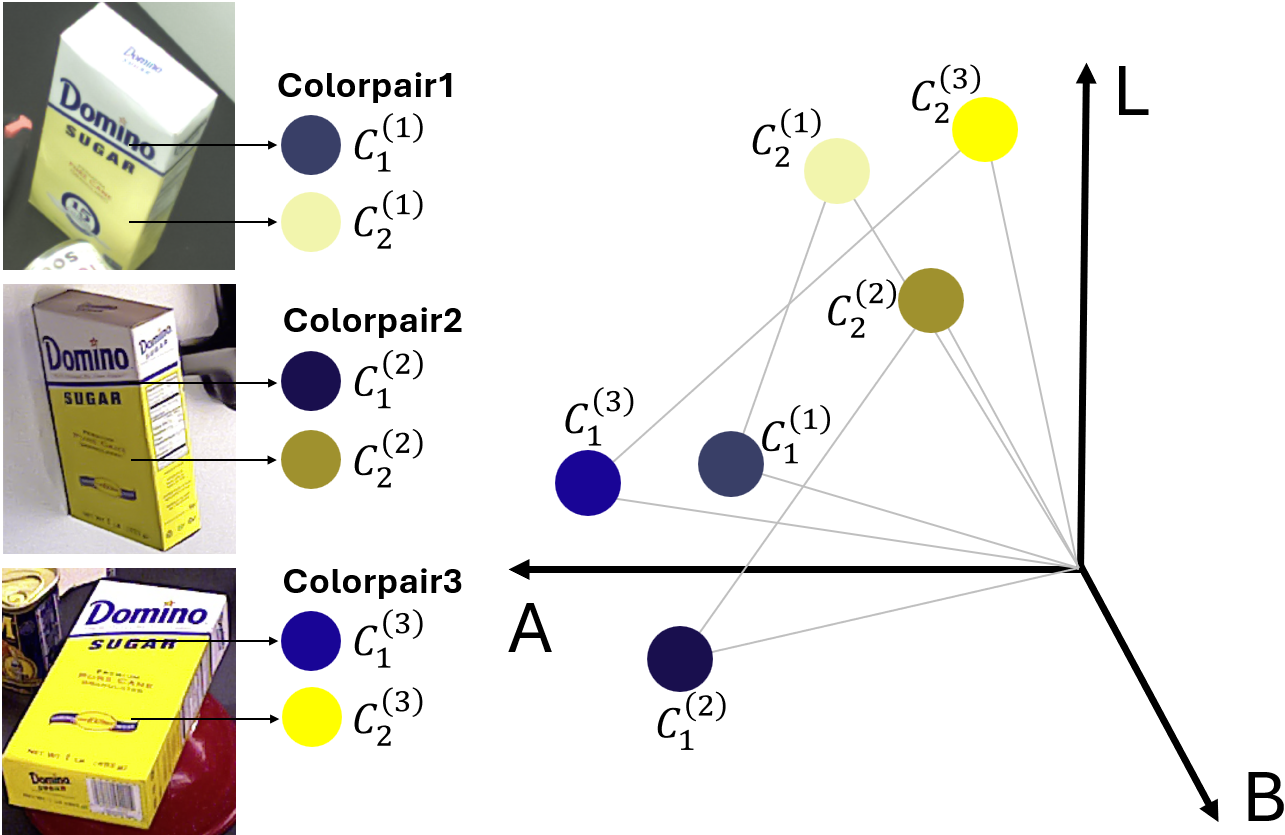}
      \caption{Color-pair consistency under illumination changes. Left: Extracted color-pairs from the same object under different lighting conditions. Right: Visualization in the CIELAB color space, where each pair with the origin forms a triangle whose pose and edge relations remain consistent across lighting variations.}
      \label{brief_illustration}
   \end{figure}

Our pipeline begins by localizing the target object in the input RGB-D frame. We first employ a YOLO detector to obtain a coarse bounding box, which is then refined by the Segment Anything Model (SAM) to yield a precise instance mask. From the masked region, we introduce our core contribution for pose estimation: a novel feature representation based on classified color pairs. By sampling colors on both sides of texture edges, we create descriptors that are robust to illumination changes. We then generate a classified edge point cloud from the image and register it against a pre-computed, similarly classified, ground truth point cloud derived from the object's 3D mesh. This registration is performed in two steps: a Hough voting algorithm provides a coarse initial pose, which is then refined using a weighted Iterative Closest Point (ICP) algorithm that jointly optimizes correspondences across all color pair classes. Once initialized, our tracking module takes over. It computes dense optical flow between consecutive frames to establish temporal correspondences. These 2D matches, augmented with 3D information from the depth sensor, are fed into a lightweight regression model that predicts the inter-frame transformation, enabling efficient and continuous 6D pose tracking. We now describe each module of our framework in detail.

\subsection{ColorPair Similarity} 
The foundation of our unified framework rests on a key observation: the relational characteristics of color pairs across local texture edges represent an intrinsic property of an object's surface that is remarkably robust to domain shifts. This principle holds true whether the object is represented as a synthetically rendered 3D model or captured in a real-world photograph under challenging illumination. This inherent, domain-invariant consistency is the very property that allows our system to bridge the gap between the offline model and online sensor data. Our colorpair Similarity metric is therefore not merely an engineered function, but a principled method designed to quantify this stable physical phenomenon, creating a consistent foundation for both initial pose estimation and subsequent tracking.

To compare two colorpairs', \(i\) and \(j\), we represent them as triangles, \(\Delta O C_1^{(i)} C_2^{(i)}\) and \(\Delta O C_1^{(j)} C_2^{(j)}\), within the LAB color space, where \(O\) is the origin (black) (see Fig. \ref{brief_illustration}). The similarity score is the product of three geometric comparisons:

\begin{itemize}
    \item \textbf{Directional Alignment:} Assesses the alignment between corresponding sides \(OC_1^{(i)}\) and \(OC_1^{(j)}\), and between \(OC_2^{(i)}\) and \(OC_2^{(j)}\). The lightness (L*) channel's contribution is down-weighted to prioritize chrominance.
    \item \textbf{Internal Contrast:} Evaluates the directional alignment between the triangles' third sides, the vectors \(C_1^{(i)}C_2^{(i)}\) and \(C_1^{(j)}C_2^{(j)}\). Its lightness component is similarly down-weighted for robustness.
    \item \textbf{Relative Luminance:} Compares the internal luminance ratio of vertices \((C_1^{(i)}, C_2^{(i)})\) to that of \((C_1^{(j)}, C_2^{(j)})\), enforcing consistent internal lighting dynamics.
\end{itemize}

Finally, to ensure the metric is robust against the arbitrary order in which colors on either side of an edge are sampled, the comparison is performed symmetrically. An observed feature from the scene is matched against both a reference feature and its color-swapped counterpart. The maximum of the two resulting scores is taken as the definitive similarity, ensuring that the matching is invariant to the sampling order. This multi-faceted function provides a highly discriminative score for reliably identifying corresponding features, transforming a fundamental physical observation into a powerful computational tool that unifies the distinct challenges of initial estimation and continuous tracking.

\subsection{Edge Color-Pair Extraction}
\textbf{Preprocessing}
To ensure reliable feature extraction, we first suppress chromatic noise using a lightweight two-stage bilateral filter. Then a joint bilateral filter on the chromatic channels, guided by luminance, further attenuates spurious color variations, resulting in a cleaner input image for subsequent processing.

\textbf{Unified Gradient Computation}
We convert the image to the CIELAB color space for robustness to illumination changes. A unified gradient is then computed where its magnitude is calculated by aggregating the energy from all three channels, while its direction is governed by the channel with the strongest local change.

\textbf{Edge Localization and Thickness Estimation}
Building on the gradient map, we apply a non-maximal suppression (NMS) procedure to distill the gradient field into crisp, one-pixel-wide contour centerlines. Our method extends NMS by concurrently measuring edge thickness. This is achieved by an iterative process where pixels within a gradient region associate with their strongest neighbor, ultimately converging to a central ridge. The number of pixels that are grouped onto each ridge point naturally defines its thickness, yielding a width map that encodes the sharpness of each contour.

\begin{figure}[b]
      \centering
      \includegraphics[scale=0.2]{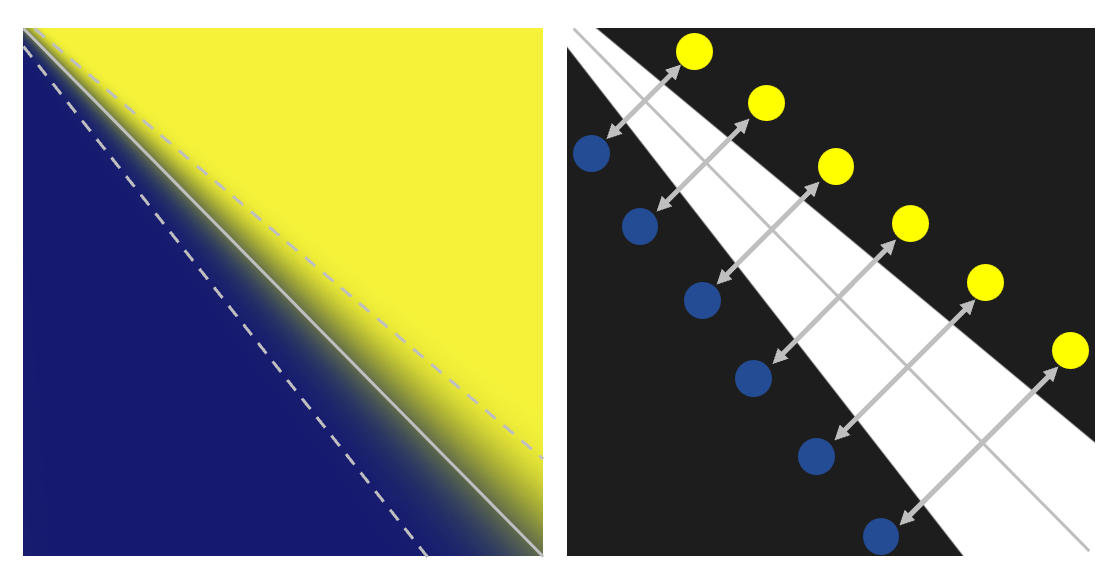}
      \caption{Thickness-aware color-pair extraction. Left: An edge image showing a diagonal centerline whose progressively increasing width is visualized by diverging dashed lines. Right: Color-pair extraction, where the width of the central white band represents the local edge thickness that guides the color sampling.}
      \label{colorpair_extraction}
   \end{figure}

\textbf{Color-Pair Sampling}
The final step is to sample representative colors from both sides of each contour centerline, guided by its direction and measured width (see Fig. \ref{colorpair_extraction}). To overcome the limitations of sampling a single point at an imprecise sub-pixel location, we first sample a profile of multiple candidates on each side of the edge at distances proportional to the local contour thickness. To robustly aggregate these candidates, we introduce a gating mechanism that operates in a statistically whitened CIELAB space. This whitening normalizes the local color distribution, leading to more stable geometric computations. Within this space, an initial color transition axis is defined, and any sampled points that are significant outliers with respect to this axis are rejected. A final representative color for each side is then computed as the median of the remaining inlier samples. This multi-stage process—combining adaptive multi-sampling with a robust gating mechanism—ensures the final color-pair accurately represents the underlying surface appearances, even across complex textures or noisy regions.

\subsection{Initial Pose Voting}
To prepare for voting, we generate a classified scene point cloud. This is achieved by first robustly matching extracted color-pairs to a reference database. High-confidence matches are then decoupled into their constituent representative colors, which in turn are used to assign a semantic class and weight to each scene point. Our approach addresses the inherent challenge of initial pose estimation in the high-dimensional SE(3) search space by reframing the continuous alignment problem into a discrete, computationally efficient geometric hashing query. This method is designed to be robust against noise, clutter, and viewpoint variations by building a consensus from numerous locally consistent geometric correspondences.

The core of our method is an offline database constructed from the reference model's point cloud. We prioritize the selection of structurally significant semantic triangles, which are chosen via a quality metric that considers both geometric stability (triangle area) and global distinctiveness (the distance of the triangle's centroid from the object's center). This process populates the database with reliable geometric anchors. For each selected triangle, we compute a 7-dimensional, pose-invariant feature key, creating a hash table that maps these canonical features to their 3D vertex coordinates. This primary approach is supplemented by Point Pair Feature (PPF) based templates to maintain robustness in scenarios with fewer than three visible object classes, as our core feature relies on triangles constructed from points across three distinct classes.

\begin{figure}[bt]
      \centering
      \includegraphics[scale=0.18]{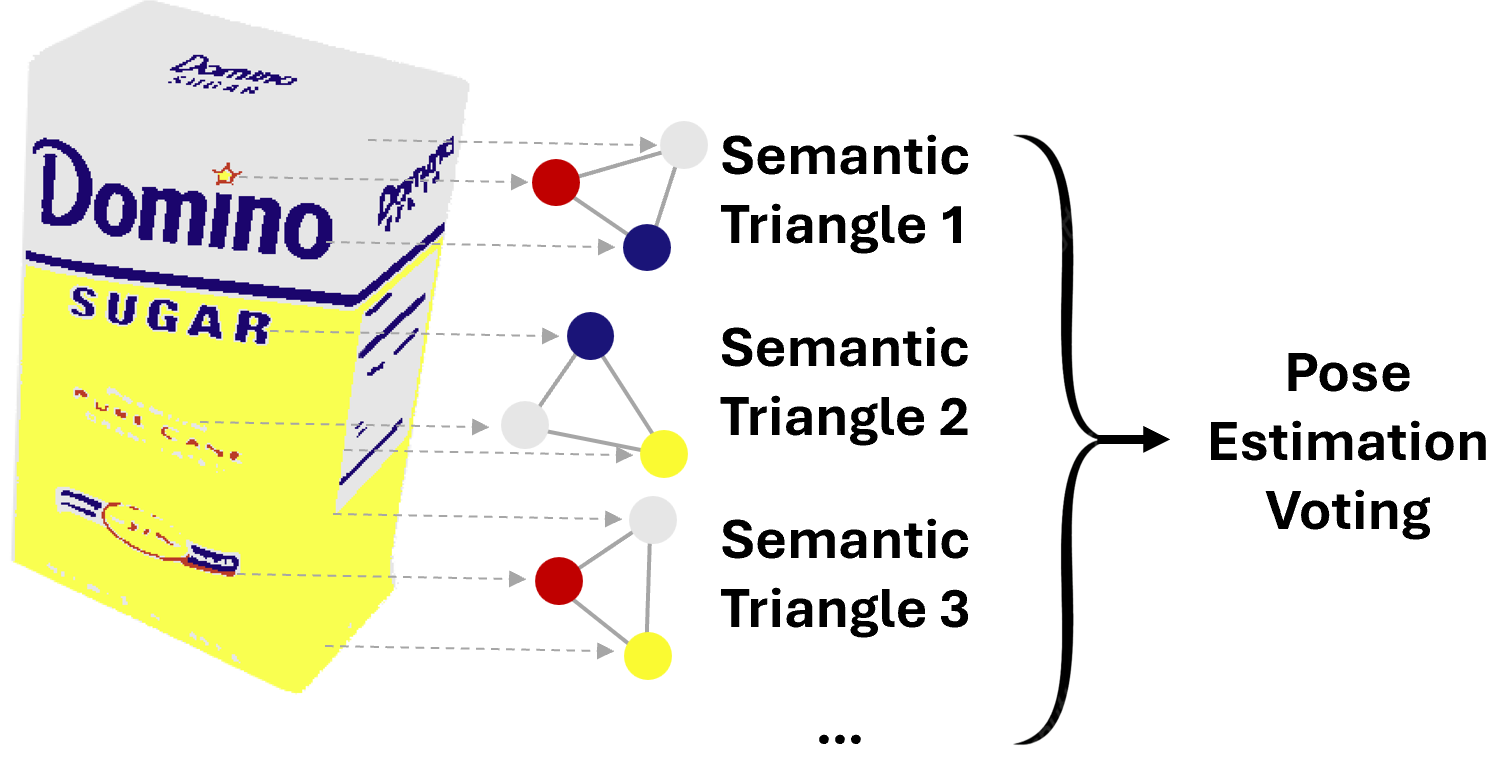}
      \caption{Illustration of the initial pose voting scheme. A classified point cloud is shown, from which several semantic triangles (1, 2, 3, …) are constructed by connecting vertices of different categories. These semantic triangles collectively contribute to the subsequent pose estimation voting process.}
      \label{pose_voting}
   \end{figure}

During the online stage, the algorithm efficiently processes a given scene point cloud by extracting local triangles, computing their feature keys, and performing a fast lookup in the pre-built database (see Fig.~\ref{pose_voting}) for an illustration). This query yields a set of candidate correspondences between scene and model triangles. For one-to-many match scenarios, where a single scene feature corresponds to multiple database entries, we employ a center-aligned equidistant sampling strategy to generate a diverse set of initial pose hypotheses using the Kabsch algorithm.

Finally, the large number of generated hypotheses, each acting as a ``weak'' estimator, are processed through a two-stage, coarse-to-fine Hough voting framework to reach a robust consensus. This framework ensures that a single scene feature does not disproportionately influence the voting outcome, thereby suppressing noise from ambiguous matches. By applying a non-maximum suppression strategy to the 6D pose space, we identify the most densely populated pose clusters. A final weighted averaging of the poses within these high-density clusters yields a small, consistent set of initial alignments suitable for subsequent fine-grained refinement.

\subsection{Pose Refinement} 
Following the initial coarse alignment, we perform a fine-grained pose refinement using a custom Iterative Closest Point (ICP) algorithm. Our method employs a composite optimization strategy that simultaneously leverages both the geometry of the individual semantic parts and the overall structure of the object within each iteration.

Specifically, the refinement process is driven by two parallel sources of geometric constraints. First, we treat the point clouds of each common semantic class (e.g., pcd1, pcd2, etc.) as distinct entities. By concurrently optimizing their alignments, we implicitly enforce the fixed spatial relationships among the object components, as they all contribute to a single, unified pose update. This preserves the high-fidelity geometric details of each part. Second, to ensure global coherence, we also form an aggregated point cloud (pcd0) by combining all individual classes. This complete point cloud, representing the overall geometric structure, is included as an additional constraint in the optimization. It acts as a holistic guide, ensuring that alignment is robust against noise, occlusion, or feature-poor regions in any part.

In every ICP iteration, correspondences from both the individual class point clouds and the aggregated point cloud are established and then jointly minimized to compute a single incremental pose update. This dual source constraint mechanism allows our method to achieve both local precision and global stability. The entire process is embedded within a multi-resolution hierarchy and uses a robust kernel-based weighting for individual point pairs to guarantee convergence and precision.

\vspace{-1.5mm}
\subsection{Pose Tracking}
Once an initial 6D pose is established, our system begins tracking the object's motion across subsequent frames. The core task is to compute the incremental transformation between the previous and current frames, allowing for a continuous and efficient pose update.

\textbf{Feature Correspondence and 3D Lifting.} We first establish robust 2D point correspondences between consecutive frames using a pre-trained optical flow network. To ensure high quality, these correspondences are filtered by propagating the object's mask and enforcing color-pair feature consistency—the same logic applied in the pose estimation section, resulting in a sparse set of reliable matches. These 2D matches are then lifted to 3D space using the camera intrinsics and corresponding depth maps, creating two point clouds in the camera's coordinate system, one for each frame.

\begin{figure}[tb]
      \centering
      \includegraphics[scale=0.2]{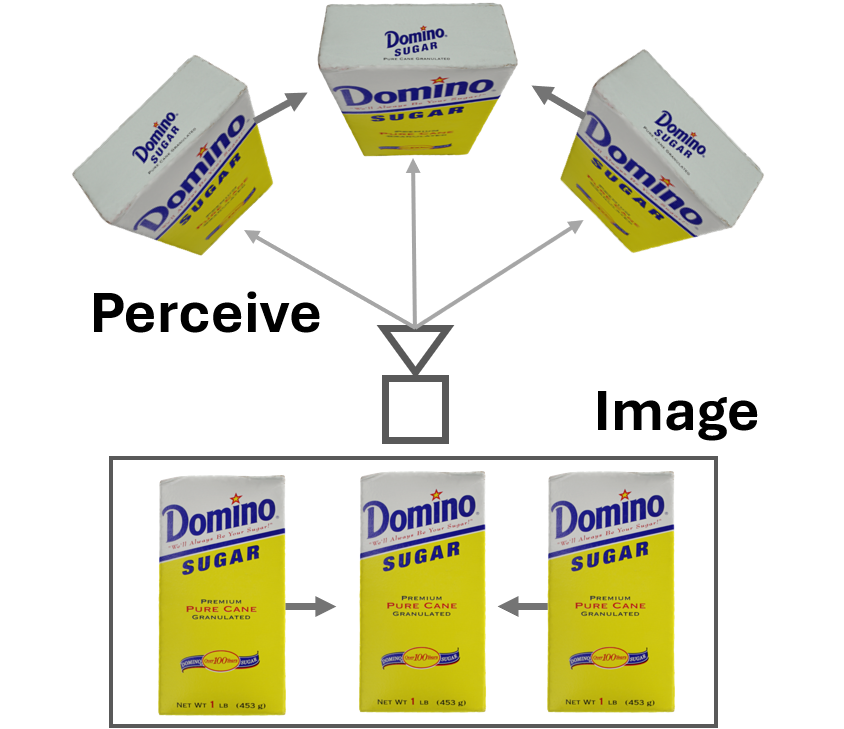}
      \caption{Illustration of Perspective Normalization. This process converts a 3D point cloud into a canonical coordinate frame to create viewpoint-invariant features. By aligning varied initial views (top) to a consistent appearance (bottom), this step decouples the object's translation from its rotation, simplifying the learning task.}
      \label{view_normalization}
   \end{figure}

\textbf{Viewpoint-Invariant Feature Generation.} A critical preprocessing step in our procedural learning approach is the generation of viewpoint-invariant features (see Fig.~\ref{view_normalization}). To ensure the model learns general principles of geometric motion rather than object-specific patterns, we first normalize both point clouds, $P_{cam,1}$ and $P_{cam,2}$, into a canonical reference frame. This normalization is a two-stage process. First, we center each point cloud by subtracting its centroid, $t_{est}$. Second, we compute a perspective correction matrix $R_p$ that reorients the point cloud into a standardized "look-at" pose. This matrix is constructed by defining a new basis where the z-axis ($z'$) aligns with the camera-to-centroid vector. The x-axis ($x'$) and y-axis ($y'$) are then determined via cross products with a predefined "up" vector, $u$ (typically $[0, -1, 0]^T$):
\begin{equation}
z' = \frac{t_{est}}{\|t_{est}\|} \quad x' = \frac{u \times z'}{\|u \times z'\|} \quad y' = z' \times x'
\end{equation}
The full transformation applied to the original point cloud $P_{cam}$ is:
\begin{equation}
P_{corrected} = R_p^T \cdot (P_{cam} - t_{est})
\end{equation}
After transforming both point clouds into this shared canonical space, we construct a 15-dimensional feature vector for each point correspondence, concatenating the normalized 3D coordinates, 2D projections, and the 3D/2D displacement flows between frames. This normalized feature set decouples the problem: by making the input independent of the object's absolute pose, the network can focus solely on regressing the relative rotation between the two canonicalized point clouds. This decoupling is key to the model's generalization across arbitrary geometries.

\textbf{Relative Pose Estimation via Procedural Learning.} The viewpoint-invariant feature vectors serve as input to our estimation model. To ensure the model generalizes to arbitrary object geometries and complex motions, we depart from traditional training on finite datasets. Instead, we introduce a novel training paradigm based on real-time procedural data generation. During training, we programmatically generate a virtually infinite stream of complex, randomized 3D shapes, each with unique topological features. For each unique shape, we simulate a realistic tracking event by rendering it from two distinct viewpoints, yielding paired, noisy point clouds with precise ground truth for the relative transformation. This methodology forces the model to learn the underlying geometric principles of motion from first principles, rather than memorizing features of specific objects.

This procedurally generated data is used to train our rotation estimator, an Attention-DGCNN that enhances the standard DGCNN \cite{wang2019dynamic} architecture by replacing its global max-pooling layer with a self-attention mechanism, which is designed to regress the relative rotation from the viewpoint-invariant features. The predicted relative rotation then serves as a strong prior for the subsequent estimation of the relative translation. Together, they form an incremental transformation that updates the object's pose. To mitigate drift from accumulated prediction errors, we refine the updated pose using a fixed small number of point-to-plane ICP iterations. This process grounds the pose by re-aligning the object's model with the current depth map, and this refined pose is then carried forward to the next tracking cycle.

\section{Experiments}

\begin{table*}[b]
\centering
\caption{Pose estimation evaluation on the YCB dataset (ADD-S / ADD).}
\label{tab:ycb_twolevel}
\resizebox{\textwidth}{!}{
\begin{tabular}{l|cc|cc|cc|cc|cc|cc|cc|cc}
\hline
Object &
\multicolumn{2}{c|}{FoundationPose\cite{wen2024foundationpose}} &
\multicolumn{2}{c|}{FreezeV2\cite{caraffa2025accurate}} &
\multicolumn{2}{c|}{GenFlow\cite{moon2024genflow}} &
\multicolumn{2}{c|}{GigaPose\cite{nguyen2024gigapose}} &
\multicolumn{2}{c|}{MegaPose\cite{labbe2022megapose}} &
\multicolumn{2}{c|}{SAM6D\cite{lin2024sam}} &
\multicolumn{2}{c|}{ZeroPose\cite{chen2024zeropose}} &
\multicolumn{2}{c}{Ours} \\
& ADD & ADD-S & ADD & ADD-S & ADD & ADD-S & ADD & ADD-S & ADD & ADD-S & ADD & ADD-S & ADD & ADD-S & ADD & ADD-S \\
\hline
002\_master\_chef\_can   & 54.1 & 84.0 & 55.8 & 82.6 & 52.9 & 81.2 & 54.1 & 81.3 & 53.5 & 76.5 & 43.0 & 75.5 & 23.9 & 78.0 & 63.0 & 77.0 \\
003\_cracker\_box        & 87.8 & 91.3 & 88.1 & 91.7 & 77.0 & 82.3 & 86.8 & 90.5 & 70.4 & 74.2 & 75.9 & 79.4 & 46.5 & 75.7 & 73.3 & 87.1 \\
004\_sugar\_box          & 81.9 & 87.8 & 90.0 & 93.8 & 78.9 & 84.6 & 83.5 & 90.0 & 72.2 & 79.1 & 88.8 & 93.1 & 72.6 & 84.8 & 73.9 & 80.3 \\
005\_tomato\_soup\_can   & 66.5 & 80.1 & 71.2 & 86.1 & 69.6 & 84.7 & 70.5 & 85.0 & 69.6 & 83.7 & 70.6 & 85.5 & 16.3 & 83.0 & 72.2 & 82.4 \\
006\_mustard\_bottle     & 87.3 & 91.6 & 91.3 & 94.0 & 77.5 & 90.3 & 79.3 & 91.2 & 76.9 & 91.2 & 90.8 & 94.1 & 62.1 & 91.5 & 77.3 & 85.3 \\
007\_tuna\_fish\_can     & 64.7 & 83.4 & 61.6 & 81.6 & 66.8 & 83.2 & 67.4 & 83.2 & 67.1 & 83.6 & 55.7 & 75.6 &  6.3 & 75.7 & 62.3 & 77.7 \\
008\_pudding\_box        & 78.3 & 87.0 & 89.1 & 92.6 & 10.8 & 22.0 & 33.8 & 43.5 & 30.9 & 46.0 & 84.1 & 89.0 & 29.7 & 59.1 & 49.3 & 74.7 \\
009\_gelatin\_box        & 81.2 & 90.2 & 90.3 & 93.0 & 77.3 & 88.6 & 77.3 & 88.6 & 77.4 & 88.3 & 87.1 & 92.0 & 76.9 & 85.2 & 78.7 & 86.7 \\
010\_potted\_meat\_can   & 59.7 & 79.8 & 60.9 & 80.1 & 51.8 & 70.4 & 61.5 & 74.9 & 57.4 & 69.4 & 46.3 & 72.2 & 28.4 & 71.8 & 60.4 & 73.8 \\
021\_bleach\_cleanser    & 73.5 & 80.8 & 86.7 & 91.3 & 71.6 & 87.5 & 72.8 & 88.0 & 76.4 & 88.6 & 87.8 & 92.9 & 56.0 & 83.3 & 79.0 & 84.3 \\
035\_power\_drill        & 85.7 & 91.3 & 88.8 & 92.6 & 82.3 & 88.1 & 84.6 & 90.5 & 73.8 & 80.6 & 85.3 & 88.4 & 81.4 & 88.7 & 71.3 & 82.3 \\
037\_scissors            & 79.6 & 85.5 & 85.2 & 92.2 & 66.9 & 81.9 & 77.3 & 89.3 & 17.5 & 34.2 & 70.4 & 81.9 & 52.8 & 83.9 & 73.3 & 78.7 \\
040\_large\_marker       & 48.1 & 88.8 & 59.2 & 90.6 & 24.5 & 89.3 & 38.3 & 89.2 & 34.3 & 90.8 & 31.2 & 87.2 & 27.2 & 86.2 & 34.0 & 82.7 \\
051\_large\_clamp        & 13.7 & 78.0 &  3.6 & 77.1 & 24.5 & 63.5 & 20.7 & 64.8 &  8.7 & 58.2 & 19.2 & 62.2 & 25.8 & 57.0 & 24.7 & 69.3 \\
052\_extra\_large\_clamp  & 19.2 & 67.8 & 24.9 & 55.9 & 21.4 & 36.4 & 25.4 & 50.5 &  8.7 & 21.6 &  9.4 & 39.9 & 13.4 & 42.7 & 20.6 & 56.7 \\
\hline
Average                   & 65.4 & 86.6 & 69.8 & 86.3 & 56.9 & 75.6 & 62.2 & 80.0 & 53.0 & 71.1 & 63.0 & 80.6 & 41.3 & 76.4 & 60.9 & 78.6 \\
\hline
\end{tabular}
}
\end{table*}


\subsection{Datasets and Setup} 
We evaluate our 6D pose estimation method on the YCB-Video (YCB-V) dataset, specifically utilizing the 12 test videos provided in the BOP format. YCB-V contains 21 objects in cluttered scenes with varying illumination. As our method relies on texture cues, we restrict our evaluation to fifteen texture-rich objects as shown in Table~\ref{tab:ycb_twolevel}. Our approach performs zero-shot pose estimation and assumes a preceding object detection stage. To simulate this input, we use the ground-truth bounding box perturbed by a random positional offset of up to 10\% of its height and width. This bounding box then serves as a prompt for the NanoSAM \cite{nvidia_nanosam} model to generate the object's segmentation mask. We report the ADD(-S) metric for evaluation, computed using the dataset’s official camera intrinsics and ground-truth poses. To compare against state-of-the-art methods, we filter their publicly available results from the BOP Challenge website for our selected object subset.

We evaluate our 6D pose tracking performance on six labeled synthetic sequences from the Fast-YCB dataset \cite{piga2021roft}, which is designed for moderate-to-fast object motions. We follow the official ROFT protocol for evaluation. All trackers are initialized using the provided DOPE pose from the first frame. Our tracker, which employs NeuFlow v2 \cite{zhang2024neuflow} for optical flow estimation, processes the main 30 Hz RGB-D stream. To simulate a resource-constrained scenario with delayed perception, the protocol stipulates that perception inputs, such as segmentation masks from Mask R-CNN, are provided at a delayed and non-synchronized 5 Hz frequency. The evaluated sequences feature the following objects: cracker box, sugar box, tomato soup can, mustard bottle, gelatin box, and potted meat can.

\begin{table*}[t]
\centering
\caption{Pose tracking evaluation on the Fast YCB dataset.}
\label{tab:ycb_crossdomain}
\resizebox{\textwidth}{!}{
\begin{tabular}{l|c|c|c|c|c|c|c}
\hline
method & DOPE\cite{tremblay2018deep} & ROFT\cite{piga2021roft} & PoseRBPF\cite{deng2021poserbpf} & SE(3)-TrackNet\cite{wen2020se} & Cross-Domain Fusion\cite{wang2025cross} & Ours & Ours (every 5 frame) \\
\hline
003\_cracker\_box   & 54.92 & 78.50 & 68.94 & 63.02 & 65.48 & 91.79 & 81.88 \\
004\_sugar\_box     & 60.01 & 81.15 & 82.78 & 73.70 & 70.88 & 96.12 & 85.76 \\
005\_tomato\_soup\_can & 64.14 & 79.00 & 75.93 & 80.82 & 77.77 & 90.50 & 75.73 \\
006\_mustard\_bottle  & 57.20 & 73.10 & 82.92 & 74.83 & 69.64 & 95.41 & 79.29 \\
009\_gelatin\_box    & 60.01 & 74.26 & 11.32 & 69.00 & 66.65 & 91.21 & 72.43 \\
010\_potted\_meat\_can & 57.03 & 73.87 & 87.29 & 71.30 & 75.06 & 93.73 & 85.43 \\
\hline
Average             & 58.83 & 76.59 & 68.10 & 72.06 & 70.91 & 93.13 & 80.09 \\
\hline
\end{tabular}
}
\end{table*}

\subsection{Evaluation and Analysis}
We evaluate 6D pose accuracy with ADD and ADD-S. Both compute the mean distance between model points transformed by the predicted and ground-truth poses—ADD uses correspondences, while ADD-S uses nearest neighbors. A pose is correct if the error is below 10\% of the object diameter.

\textbf{Pose Estimation.} We compare our method against several state-of-the-art methods including FoundationPose\cite{wen2024foundationpose}, FoundPose\cite{ornek2024foundpose}, FreeZeV2\cite{caraffa2025accurate}, GenFlow\cite{moon2024genflow}, and GigaPose\cite{nguyen2024gigapose} on the YCB-Video dataset. As shown in Table \ref{tab:ycb_twolevel}, our method achieves an average ADD-S of 78.6 and an average ADD of 60.9. Although methods such as FoundationPose (92.3 ADD-S) or FreeZeV2 (86.3 ADD-S) achieve higher accuracy scores, our approach remains competitive with top-performing methods, indicating a favorable balance between accuracy and robustness.

In terms of efficiency, these competing methods generally rely on powerful GPUs (e.g., A100 or V100) and employ a two-stage pipeline. The initial zero-shot segmentation step, often performed by models like CNOS, introduces a variable overhead: it requires approximately 0.2-0.4 seconds per frame with an efficient FastSAM \cite{zhao2023fast} model, which increases to over 1.5 seconds using the original SAM \cite{kirillov2023segment}. This overhead contributes to total reported processing times that often span multiple seconds (e.g., FoundationPose at 9.9s; GenFlow at 16.7s; FreeZeV2 at 14.3s). In contrast, our method assumes input from a fast object detector (e.g., YOLO) and is designed for on-device deployment. It achieves 7 FPS on a Jetson AGX Orin while processing five objects simultaneously, demonstrating its suitability for embedded, real-time applications with further optimization.

This distinction highlight a different design focus: while not surpassing the top methods in raw accuracy, our approach provides a pragmatic alternative, delivering competitive performance at a substantially lower computational cost. This makes it particularly well-suited for robotic applications where low latency and onboard deployment are critical.

    \begin{figure}[thpb]
      \centering
      \includegraphics[scale=0.25]{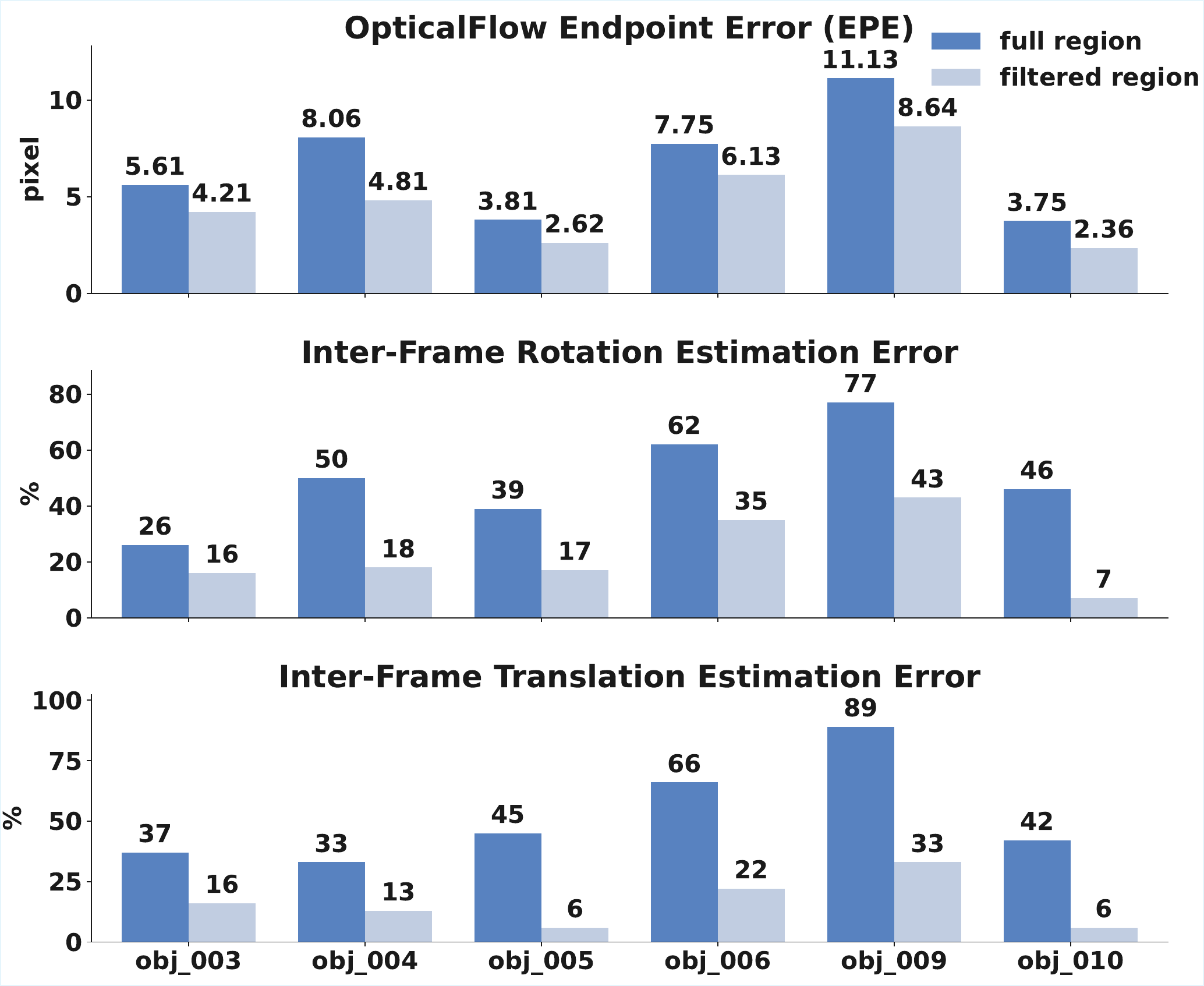}
      \caption{Quantitative comparison of Translation Error, Rotation Error, and Endpoint Error (EPE) across different objects. Translation Error denotes the magnitude of translational residuals normalized by the true translational motion. Rotation Error is the residual rotation angle normalized by the true rotational motion. EPE represents the average 2D endpoint error of predicted scene flow. Results compare pose estimated from the flow of points sampled on the full object surface versus pose estimated from the flow of points sampled on color-pair–filtered surface regions.}
      \label{pose_tracking_ablation}
   \end{figure}

\textbf{Pose Tracking.} As detailed in Table II, our proposed method demonstrates a substantial improvement in performance over all baseline approaches on the Fast-YCB dataset, which features objects undergoing rapid motion. Specifically, our approach achieves an average ADD score of 93.13. In addition to its superior accuracy, our method is highly efficient, capable of processing at over 20 FPS on the NVIDIA Jetson Orin platform, underscoring its suitability for real-world robotic applications.

To further evaluate the robustness of our tracker, we conducted an experiment where the input is downsampled to every 5th frame, effectively simulating scenarios with abrupt pose changes. Remarkably, even under this challenging condition, our method achieves an average ADD score of 84.66. This result not only demonstrates strong robustness to sudden displacements, but also surpasses the performance of all competing methods that operate on the full, dense video sequence. These findings collectively validate that our approach offers a highly accurate, efficient, and robust solution for 6D object pose tracking.

\textbf{Ablation Study.} We conduct an ablation study to validate our core contribution: using a color-pair similarity metric to filter correspondences for robust pose estimation. To isolate the quality of the resulting matches, we estimate the relative pose between frame pairs sampled at a five-frame interval using a raw estimator without any subsequent ICP refinement. This setup stresses the ability to find geometrically consistent correspondences under significant motion. We compare the pose accuracy derived from two inputs: (1) all available optical flow correspondences, and (2) only those correspondences deemed consistent by our color-pair metric. As shown in Fig.~\ref{pose_tracking_ablation}, while the optical flow's endpoint error (EPE) is only modestly reduced, the impact on the downstream pose calculation is substantial. We observe a pronounced reduction in both rotation error and translation error, with both metrics normalized by the magnitude of the ground-truth inter-frame motion. This result demonstrates that our color-pair metric is effective at rejecting ambiguous temporal matches, thereby improving the geometric integrity of the correspondences used for pose estimation.

\subsection{Proof-of-concept Demonstration}
\begin{figure}[t]
      \centering
      \includegraphics[scale=0.12]{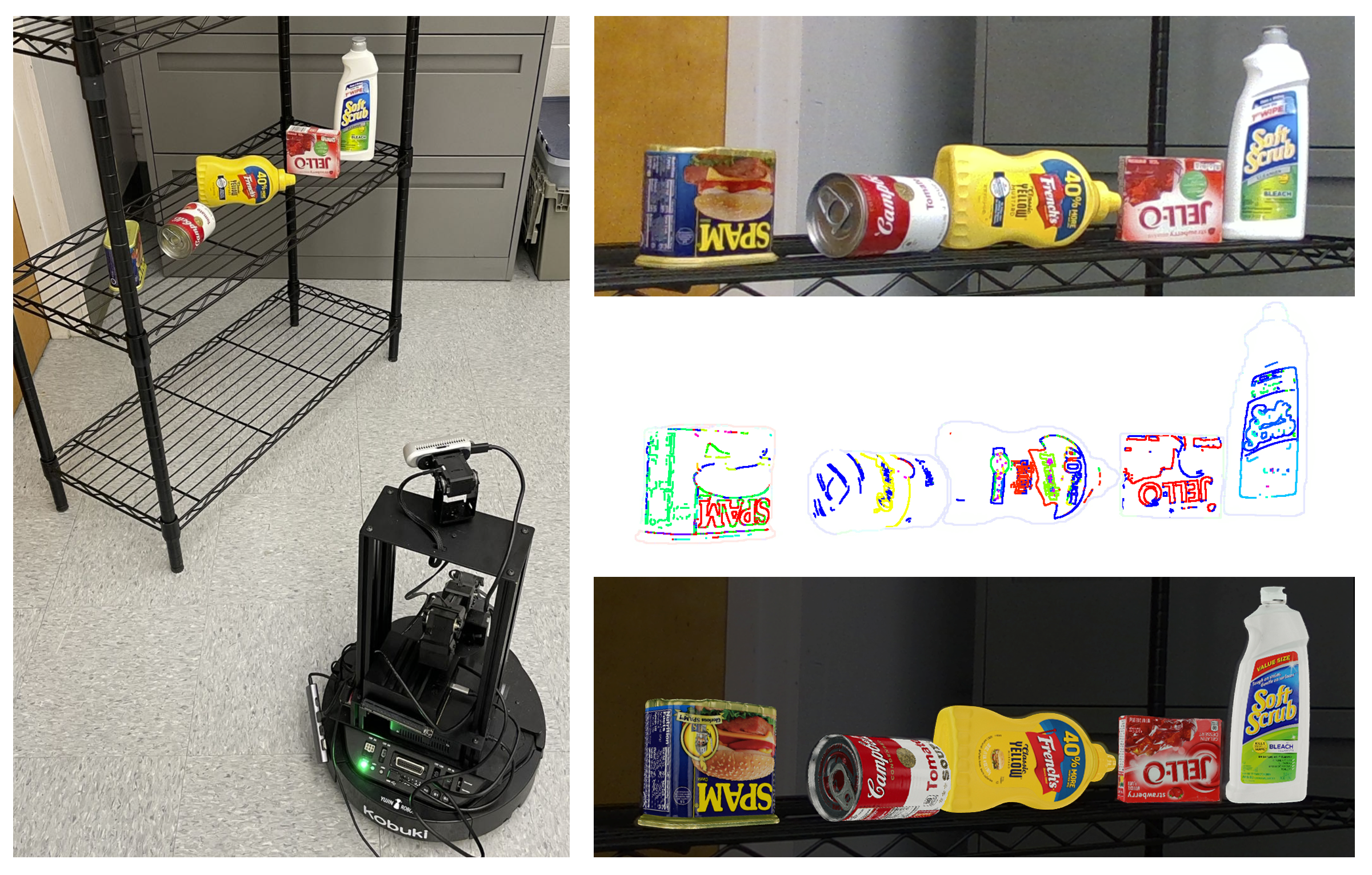}
      \caption{Real-World Experiment. Left: experimental setup. Right (top to bottom): input scene, color-pair classification, and the final rendered pose estimate.}
      \label{proof-of-concept}
   \end{figure}

Our TensorRT-optimized pipeline runs on a LoCoBot platform equipped with an Intel RealSense D435 camera. The pose estimation module processes images on an on-board NVIDIA Jetson AGX Xavier. Fig.~\ref{proof-of-concept} shows the experimental setup and a qualitative result, where the object's textured mesh is rendered at its estimated pose, demonstrating robustness to a slight texture mismatch between the physical object and its reference mesh file.

\section{Discussion}

For pose estimation, our results are competitive but do not surpass the highest-performing methods. We attribute this gap partly to our method's reliance on depth data for both voting and ICP refinement. Real-world depth maps can suffer from noise and distortions, whereas 2D positional cues from RGB images are often more precise. Future work could therefore focus on integrating these 2D cues during refinement to improve accuracy. 

Regarding pose tracking, our method demonstrates strong performance on the Fast-YCB dataset, a result attributed to the synergy of our robust filtering and refinement strategy and the dataset's clean depth data. This combination allows the tracker to operate reliably without frequent re-correction. However, the noticeable drop in accuracy when tracking at a 5-frame interval (which occasionally requires re-initialization) highlights that abrupt pose changes remain a challenge. Therefore, future work could extend the system's reliance on a single previous frame to incorporate multiple keyframes, establishing a more stable temporal baseline and improving resilience to rapid motions or feature scarcity.

\section{Conclusions}
In this paper, we present a unified framework for zero-shot 6D object pose estimation and tracking, designed for robust performance on edge devices. Our method introduces a novel, lighting-invariant edge color-pair feature for accurate initial pose estimation, which is then seamlessly integrated with a high-speed, optical flow-based tracker. Experiments demonstrate that our approach achieves competitive accuracy on standard benchmarks while operating in real-time on resource-constrained hardware, highlighting its practical value for real-world robotic applications.

\section*{Acknowledgment}
We acknowledge the use of ChatGPT in text editing.

\bibliographystyle{IEEEtran}
\bibliography{refs}

\end{document}